\documentclass[conference]{IEEEtran}
\IEEEoverridecommandlockouts
\usepackage{cite}
\usepackage{amsmath,amssymb,amsfonts}
\usepackage{algorithmic}
\usepackage{graphicx}
\usepackage{textcomp}
\usepackage{xcolor}
\usepackage{dm-colors}
\usepackage[pagebackref=false,breaklinks=false,%
            colorlinks=true,bookmarks=true,citecolor=ourdarkblue,%
            urlcolor=ourdarkblue,linkcolor=ourdarkblue]{hyperref}

\usepackage{comment}
\usepackage{tcolorbox}
\usepackage{tabu}
\usepackage{multirow}

\tcbuselibrary{breakable}

\def\BibTeX{{\rm B\kern-.05em{\sc i\kern-.025em b}\kern-.08em
    T\kern-.1667em\lower.7ex\hbox{E}\kern-.125emX}}
    
\begin{document}

\renewcommand\arraystretch{1.5}

\title{Development of a Large Language Model-based Multi-Agent Clinical Decision Support System for Korean Triage and Acuity Scale (KTAS)-Based Triage and Treatment Planning in Emergency Departments\\
}

\author{\IEEEauthorblockN{1\textsuperscript{st} Seungjun Han}
\IEEEauthorblockA{
\textit{John Bapst Memorial High School}\\
Bangor, USA \\
henryhansj@gmail.com}
\and
\IEEEauthorblockN{2\textsuperscript{nd} Wongyung Choi}
\IEEEauthorblockA{
\textit{Minerva University}\\
San Francisco, USA \\
woncho@uni.minerva.edu}
}

\maketitle

\begin{abstract}
Emergency department (ED) overcrowding and the complexity of rapid decision-making in critical care settings pose significant challenges to healthcare systems worldwide. While clinical decision support systems (CDSS) have shown promise, the integration of large language models (LLMs) offers new possibilities for enhancing triage accuracy and clinical decision-making. This study presents an LLM-driven CDSS designed to assist ED physicians and nurses in patient triage, treatment planning, and overall emergency care management.

We developed a multi-agent CDSS utilizing Llama-3-70b as the base LLM, orchestrated by CrewAI and Langchain. The system comprises four AI agents emulating key ED roles: Triage Nurse, Emergency Physician, Pharmacist, and ED Coordinator. It incorporates the Korean Triage and Acuity Scale (KTAS) for triage assessment and integrates with the RxNorm API for medication management.

The model was evaluated using the Asclepius dataset, with performance assessed by a clinical emergency medicine specialist. The CDSS demonstrated high accuracy in triage decision-making compared to the baseline of a single-agent system. Furthermore, the system exhibited strong performance in critical areas, including primary diagnosis, critical findings identification, disposition decision-making, treatment planning, and resource allocation.

Our multi-agent CDSS demonstrates significant potential for supporting comprehensive emergency care management. By leveraging state-of-the-art AI technologies, this system offers a scalable and adaptable tool that could enhance emergency medical care delivery, potentially alleviating ED overcrowding and improving patient outcomes. This work contributes to the growing field of AI applications in emergency medicine and offers a promising direction for future research and clinical implementation.
\end{abstract}

\begin{IEEEkeywords}
emergency department overcrowding, clinical decision support system, large language model, multi-agent system
\end{IEEEkeywords}

\section{Introduction}
Emergency departments (EDs) worldwide are at the forefront of healthcare, serving as critical access points for urgent medical care. However, these vital facilities face unprecedented challenges that threaten their ability to provide timely and effective care. The most pressing of these challenges is overcrowding, a phenomenon that has reached crisis levels in many countries. For instance, ED visits increased by 23.1\% in the United States from 1997 to 2007, reaching 116.8 million annual visits \cite{b1}. Similar trends have been observed globally, including in South Korea, where the number of patients who were treated in EDs in Korea was 10,609,107, which was an increase of 1.76\% compared to the previous year 2017.\cite{b2,b3}.

The consequences of ED overcrowding are severe and multifaceted. Prolonged wait times not only cause patient discomfort but can also lead to delayed treatments, potentially exacerbating medical conditions\cite{b19}. It is evident that ED waiting times are associated with short-term mortality and hospital admission in high-acuity patients. Furthermore, overcrowding has been linked to increased medical errors, reporting that ED crowding was associated with increased adverse events, medication errors, and patient mortality\cite{b4,b30}.

The strain on healthcare providers in overcrowded EDs is equally concerning. Emergency physicians and nurses work under immense pressure to make rapid decisions with often incomplete information. This high-stress environment contributes to burnout among healthcare professionals, a problem that has been exacerbated in recent years\cite{b5}. ED physicians experiencing burnout symptoms has been a serious problem, significantly higher than the average across all physician specialties\cite{b6}.

In this challenging environment, the need for efficient triage and rapid yet accurate clinical decision-making is paramount. Triage, the process of prioritizing patients based on the severity of their condition, is particularly crucial. The Korean Triage and Acuity Scale (KTAS)\cite{b15}, adapted from the Canadian Triage and Acuity Scale\cite{b20}, has been widely adopted in South Korea for standardizing emergency triage processes. While effective, the implementation of KTAS still faces challenges related to inter-rater reliability and the cognitive load placed on triage nurses, especially during peak hours. There were also discrepancies in triage levels assigned by different nurses, particularly for mid-acuity patients\cite{b7}.

To address these challenges, healthcare systems have increasingly turned to technology-driven solutions\cite{b31}. Clinical Decision Support Systems (CDSS)\cite{b33} has emerged as a promising approach to enhancing clinical decision-making and improving patient outcomes. Traditional CDSS, often rule-based or utilizing simple machine learning algorithms, have shown potential in improving clinical outcomes and reducing medical errors. For instance, Ehteshami et al. revealed that health information technologies, including CDSS, improved adherence to clinical guidelines and reduced medication errors in various healthcare settings\cite{b8}.

However, the application of CDSS in emergency medicine presents unique challenges. The dynamic and unpredictable nature of emergency care, coupled with the need for rapid decision-making, often stretches the capabilities of traditional CDSS. These systems frequently struggle with the complexity and variability inherent in emergency medicine, leading to limited adoption and impact\cite{b23,b24}. 

Recent advancements in artificial intelligence, particularly in the field of large language models (LLMs), offer new possibilities for enhancing CDSS capabilities. LLMs, such as GPT-4 and its successors\cite{b25}, have demonstrated remarkable proficiency in understanding context, processing natural language, and generating human-like responses across various domains. These models also have shown promise in medical applications\cite{b26}, including clinical text summarization, medical question answering\cite{b27}, and even simulating patient-doctor conversations\cite{b28}.

Despite these advancements, the application of LLMs in emergency medicine remains largely unexplored, presenting a significant research opportunity. The potential of LLMs to understand and process complex medical information, combined with their ability to generate contextually appropriate responses, makes them particularly well-suited for the fast-paced, high-stakes environment of the ED.

In this study, we seek to address this gap by proposing a novel multi-agent CDSS powered by state-of-the-art large language models. This system aims to integrate KTAS-based triage with comprehensive emergency care management, leveraging advanced AI technologies to support emergency room physicians and nurses in their decision-making processes.

\section{Materials and Methods}

\subsection{System Architecture}

\begin{figure*}[htbp]
\begin{centering}
\centerline{\includegraphics[width=\textwidth]{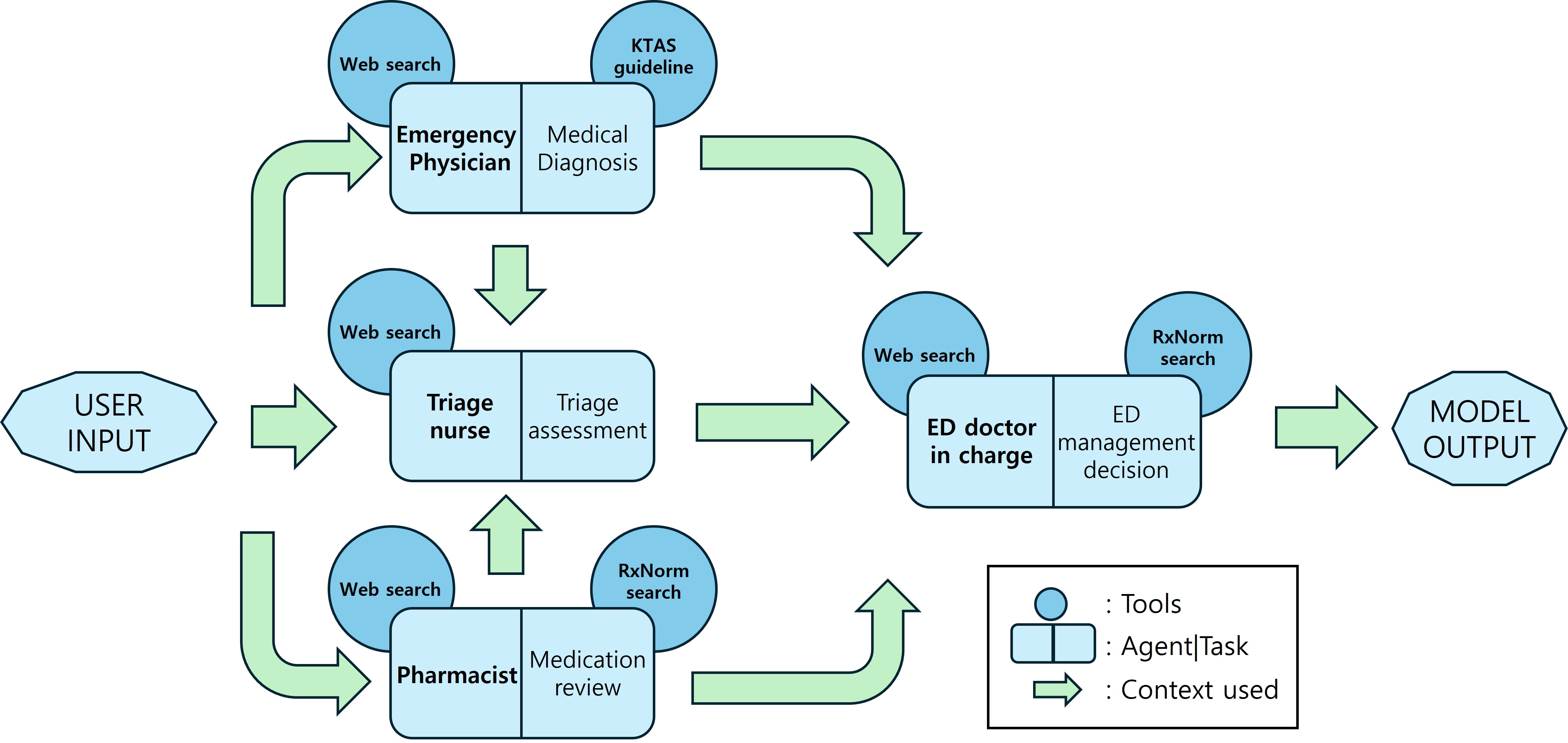}}
\caption{Model architecture.}
\label{fig1}
\end{centering}
\end{figure*}

Our Clinical Decision Support System (CDSS) is built upon a multi-agent architecture powered by the Llama-3-70b large language model\cite{b13}.
At its core, Llama-3-70b serves as the foundation, chosen for its open-source capability and advanced natural language understanding. 
The system's multi-agent framework is orchestrated by CrewAI, which enables the creation and management of multiple AI agents \cite{b14}. This framework allows us to simulate key emergency room roles, enhancing the system's ability to provide comprehensive and coordinated care recommendations. Each agent operates with a specific role, goal, and set of tools, mirroring the collaborative nature of real emergency department teams. Fig.~\ref{fig1} provides a schematic representation of our model's workflow.

\subsection{AI Agents}

Our CDSS comprises four distinct AI agents, each emulating a crucial role in the emergency department. The Emergency Physician Agent initiates the process by diagnosing patients and developing treatment plans. This agent analyzes the patient's presenting information along with additional examination data to formulate a diagnosis and differential diagnoses, consider potential complications, and outline necessary follow-up care. Access to up-to-date medical information through an integrated search function enhances the agent's capabilities, allowing it to base its recommendations on the latest evidence-based practices.

Concurrently, medication safety is ensured by the Pharmacist Agent, who reviews the patient's medication history, checks for drug interactions, and provides crucial information about proper medication administration and potential side effects. This agent leverages the RxNorm dataset \cite{b16}, a comprehensive source of standard names for clinical drugs, to verify medications, check for interactions, and ensure appropriate dosing.

Following these initial assessments, the Triage Nurse Agent conducts the patient assessment using the Korean Triage and Acuity Scale (KTAS)\cite{b15}. This agent synthesizes the information provided by the Emergency Physician and Pharmacist agents, along with the patient's symptoms, vital signs, and medical history, to determine the urgency of care needed. The agent's decision-making process is guided by KTAS guidelines, ensuring accurate and consistent triage classifications based on a comprehensive view of the patient's condition.

Finally, the ED Doctor in Charge agent oversees the entire patient care process. This agent integrates information from all other agents to make critical decisions about patient management, including finalizing treatment plans, determining admission or discharge, and planning follow-up care. The agent's role is to ensure coherent and comprehensive patient management, taking into account both clinical needs and resource utilization based on the collective insights of all preceding agents.

\subsection{Data Integration and Tool Utilization}

We integrated several key data sources and tools to enhance the system's decision-making capabilities. The KTAS guidelines were implemented as a comprehensive prompt, enabling the Triage Nurse Agent to accurately classify patients based on their presenting symptoms and vital signs. The RxNorm dataset (version 2023AB) serves as the backbone for the Pharmacist Agent's medication management capabilities, providing a standardized nomenclature for clinical drugs and facilitating accurate interaction checking and dosing recommendations. Every agent has the ability to search the web via the DuckDuckGo search tool\cite{b29}. 

\subsection{System Workflow}

The system employs a streamlined workflow that mirrors the collaborative nature of emergency department operations while optimizing patient care. At the core of this system, the Emergency Physician Agent initiates the process by analyzing the patient's presenting information to generate an initial diagnosis and treatment strategy. Simultaneously, the Pharmacist Agent conducts a thorough review of the patient's medication history, utilizing the RxNorm database to identify potential drug interactions and provide critical pharmacological insights. These preliminary assessments inform the subsequent actions of the Triage Nurse Agent, who performs a comprehensive evaluation of the patient's condition, assigns a KTAS level, and compiles a succinct yet detailed triage report. The workflow culminates with the ED Doctor in Charge Agent, which integrates all preceding analyses to formulate definitive care decisions and management strategies.

\subsection{Evaluation Methodology}

We evaluated our CDSS using the Asclepius dataset, which comprises simulated emergency department scenarios\cite{b17}. Utilizing randomly selected 43 cases, scenarios were carefully curated to represent a diverse range of emergency medical situations, providing a comprehensive test of the system's capabilities across various clinical presentations.

An experienced emergency physician reviewed the CDSS outputs for each scenario. He assessed the accuracy of KTAS classification, appropriateness of diagnosis and treatment recommendations, comprehensiveness of medication reviews, and overall clinical decision quality. This expert evaluation provides a clinically relevant assessment of the system's performance.

To quantify the benefits of our multi-agent approach, we conducted a comparative analysis against a single-agent system using Llama-3-70b alone, with ED Doctor in Charge agent prompt. Key performance metrics included the concordance rate with expert KTAS classification, a clinical appropriateness score for management recommendations, and a coherence improvement rate compared to the single-agent system.

\section{Results}

\subsection{Performance of Triage Assessment with KTAS}

We evaluated the performance of our CDSS in classifying patients according to KTAS levels. The system's performance was compared to that of a single-agent system, which served as a control. Both systems were evaluated against the KTAS levels assigned by an experienced ED physician. Table \ref{tab:multi-agent-ktas} and Table \ref{tab:single-agent-ktas} show the results of KTAS assessment evaluation for multi-agent and single-agent systems, respectively.

\begin{table}[htbp]
\caption{Confusion Matrix of KTAS Classification by our CDSS}
\begin{center}
\begin{tabu}{ | c | X[c] | X[c] | X[c] | X[c] | X[c]|}
\hline
\multirow{2}{*}{\textbf{Model's KTAS Prediction}} & \multicolumn{5}{|c|}{\textbf{KTAS Level by ED Physician}} \\
\cline{2-6}
& \textbf{1} & \textbf{2} & \textbf{3} & \textbf{4} & \textbf{5} \\
\hline
\textbf{1} & 3 & 0 & 0 & 0 & 0 \\
\hline
\textbf{2} & 0 & 15 & 2 & 2 & 1 \\
\hline
\textbf{3} & 0 & 1 & 7 & 5 & 1 \\
\hline
\textbf{4} & 0 & 0 & 0 & 0 & 0 \\
\hline
\textbf{5} & 0 & 0 & 0 & 0 & 5 \\
\hline
\end{tabu}
\label{tab:multi-agent-ktas}
\end{center}
\end{table}

\begin{table}[htbp]
\caption{Confusion Matrix of KTAS Classification by Single-agent System}
\begin{center}
\begin{tabu}{ | c | X[c] | X[c] | X[c] | X[c] | X[c]|}
\hline
\multirow{2}{*}{\textbf{Model's KTAS Prediction}} & \multicolumn{5}{|c|}{\textbf{KTAS Level by ED Physician}} \\
\cline{2-6}
& \textbf{1} & \textbf{2} & \textbf{3} & \textbf{4} & \textbf{5} \\
\hline
\textbf{1} & 2 & 1 & 0 & 0 & 0 \\
\hline
\textbf{1 or 2} & 0 & 4 & 1 & 0 & 0 \\
\hline
\textbf{2} & 0 & 8 & 3 & 1 & 2 \\
\hline
\textbf{3} & 0 & 2 & 5 & 4 & 3 \\
\hline
\textbf{3 or 4} & 0 & 0 & 0 & 0 & 1 \\
\hline
\textbf{4} & 0 & 0 & 0 & 0 & 0 \\
\hline
\textbf{5} & 0 & 0 & 0 & 0 & 0 \\
\hline
\textbf{Not applicable} & 1 & 1 & 0 & 2 & 1 \\
\hline
\end{tabu}
\label{tab:single-agent-ktas}
\end{center}
\end{table}

The multi-agent system accurately classified KTAS level 1 (most urgent), level 2, and level 5 (least urgent) cases. The system showed good performance, especially for KTAS level 2, which had a small number of misclassifications. However, the system encountered some challenges in differentiating between KTAS levels 3 and 5, suggesting an area for potential refinement.

In contrast, the single-agent system exhibited lower overall accuracy compared to the multi-agent system. A notable characteristic of the single-agent system was its tendency to provide range classifications (e.g., "1 or 2", "3 or 4") rather than specific KTAS levels. Moreover, in some cases, the single-agent system failed to predict any KTAS level, resulting in non-applicable (N/A) outputs.

The multi-agent approach demonstrated superior performance in several key aspects compared to the single-agent system. The multi-agent system consistently provided definitive KTAS classifications, while the single-agent system often resorted to range predictions or even did not predict the level (not applicable). Moreover, the multi-agent system exhibited higher overall accuracy, particularly for KTAS levels 1, 2, and 5. The multi-agent system's predictions also showed greater consistency with the ED physician's evaluations across all KTAS levels.

\subsection{Evaluation on Clinical Decision Making}

We evaluated the clinical decision-making capabilities of both the multi-agent and single-agent systems across various categories crucial to emergency care. The evaluation was conducted using two scales: a 5-point scale (1: very inaccurate, 5: very accurate) for primary diagnosis, critical findings, disposition decision, and justification; and a 1-point scale (0: inaccurate, 0.5: moderate, 1: accurate) for immediate action, medication, diagnostic test, consultation, and monitoring.

\subsubsection{5-Point Scale Evaluation}
Table \ref{tab:5-point-scale} presents the results of the 5-point scale evaluation for both systems.
\begin{table}[htbp]
\caption{5-Point Scale Evaluation of Clinical Decision Making}
\begin{center}
\begin{tabu}{|l|X[c]|X[c]|X[c]|X[c]|X[c]|X[c]|X[c]|X[c]|X[c]|X[c]|}
\hline
\multirow{2}{*}{\textbf{Category}} & \multicolumn{5}{c|}{\textbf{Multi-agent System}} & \multicolumn{5}{c|}{\textbf{Single-agent System}} \\
\cline{2-11}
& \textbf{1} & \textbf{2} & \textbf{3} & \textbf{4} & \textbf{5} & \textbf{1} & \textbf{2} & \textbf{3} & \textbf{4} & \textbf{5} \\
\hline
Primary Diagnosis & 0 & 0 & 0 & 2 & 41 & 0 & 0 & 0 & 2 & 41 \\
\hline
Critical Findings & 0 & 0 & 0 & 1 & 42 & 0 & 0 & 3 & 0 & 40 \\
\hline
Justification & 0 & 0 & 0 & 0 & 43 & 0 & 0 & 2 & 4 & 37 \\
\hline
\end{tabu}
\label{tab:5-point-scale}
\end{center}
\end{table}

Both systems demonstrated high accuracy in primary diagnosis, with identical performance. However, the multi-agent system showed superior performance in identifying critical findings and providing justifications, with a higher proportion of cases rated as "very accurate" (5 points).

\subsubsection{1-Point Scale Evaluation}
Table \ref{tab:1-point-scale} shows the results of the 1-point scale evaluation for both systems.
\begin{table}[htbp]
\caption{1-Point Scale Evaluation of Clinical Decision Making}
\begin{center}
\begin{tabu}{|l|X[c]|X[c]|X[c]|X[c]|X[c]|X[c]|}
\hline
\multirow{2}{*}{\textbf{Category}} & \multicolumn{3}{c|}{\textbf{Multi-agent System}} & \multicolumn{3}{c|}{\textbf{Single-agent System}} \\
\cline{2-7}
& \textbf{0} & \textbf{0.5} & \textbf{1} & \textbf{0} & \textbf{0.5} & \textbf{1} \\
\hline
Disposition Decision & 0 & 1 & 42 & 4 & 4 & 35 \\
\hline
Immediate Action & 0 & 0 & 43 & 1 & 4 & 38 \\
\hline
Medication & 0 & 0 & 43 & 0 & 3 & 40 \\
\hline
Diagnostic Test & 0 & 0 & 43 & 0 & 0 & 43 \\
\hline
Consultation & 0 & 0 & 43 & 1 & 0 & 42 \\
\hline
Monitoring & 0 & 0 & 43 & 0 & 0 & 43 \\
\hline
\end{tabu}
\label{tab:1-point-scale}
\end{center}
\end{table}

The multi-agent system consistently outperformed the single-agent system across all categories evaluated on the 1-point scale. Notably, the multi-agent system achieved perfect accuracy in immediate action, medication, diagnostic tests, consultations, and monitoring. While performing well, the single-agent system showed some inaccuracies and moderate performances across these categories.

These results indicate that the multi-agent system provides more consistent and accurate clinical decision-making support across various aspects of emergency care. The system's superior performance in critical areas suggests its potential to enhance the quality and safety of emergency care.

\section{Discussion}

\subsection{Model performance}

Our LLM-based multi-agent CDSS demonstrated superior performance compared to the single-agent system across multiple dimensions of emergency care decision-making. The system's effectiveness was particularly evident in KTAS-based triage and clinical decision-making.

Our multi-agent system exhibited high accuracy in KTAS classification, especially for the most urgent (KTAS level 1) and urgent (KTAS level 2) cases. This level of accuracy is crucial in emergency settings, where rapid and correct triage can significantly impact patient outcomes. 

However, we observed a tendency for the system to overestimate urgency, particularly for KTAS levels 3 to 5. This bias towards higher urgency ratings contrasts with the typically more conservative approach of human ED physicians, who are trained to err on the side of caution. This discrepancy highlights an important difference between the decision-making processes of our LLM-based system and human clinicians. While the system's cautious approach might lead to over-triage in some cases, it could potentially reduce the risk of overlooking critically ill patients.

The multi-agent system's superiority over the single-agent system was particularly apparent in its decisiveness and consistency. While the single-agent system often provided range classifications or failed to predict a KTAS level, our multi-agent system consistently offered definitive classifications. This decisiveness is invaluable in real-world emergency departments where clear, rapid decisions are essential for efficient patient flow and resource allocation.

In terms of clinical decision-making, the multi-agent system demonstrated high accuracy across all evaluated categories. Notably, it achieved perfect scores in immediate action, medication, diagnostic tests, consultation, and monitoring on the 1-point scale evaluation. The system's strong performance in identifying critical findings and providing justifications for decisions, as evidenced by the 5-point scale evaluation, suggests its potential to enhance the quality and safety of emergency care delivery.

The consistent outperformance of the multi-agent system over the single-agent system across various aspects of emergency care decision-making underscores the effectiveness of our approach. This performance difference likely stems from the multi-agent system's ability to mimic the collaborative nature of real emergency department teams, where different specialists contribute their expertise to patient care.

\subsection{Mutli-agent System Effectiveness}

The superior performance of our multi-agent CDSS compared to the single-agent system underscores the potential of multi-agent strategies in leveraging LLMs for healthcare applications. This approach represents a significant advancement in the field of clinical decision support, particularly in the complex and time-sensitive environment of emergency departments.

Our multi-agent system's effectiveness stems from its ability to emulate the collaborative nature of real emergency department teams. By dividing tasks among specialized agents - the Triage Nurse, Emergency Physician, Pharmacist, and ED Doctor in Charge - the system mirrors the diverse expertise found in actual ED settings. This division of labor allows each agent to focus on its specific domain, leading to more accurate and comprehensive decision-making.

The Triage Nurse agent's proficiency in KTAS classification, coupled with the Emergency Physician agent's diagnostic capabilities, creates a robust initial assessment process. The Pharmacist agent's role in medication management adds a crucial layer of patient safety, while the ED Doctor in Charge agent synthesizes all inputs to make final decisions. This structure enables the system to handle the multifaceted nature of emergency care more effectively than a single-agent approach.

Moreover, the multi-agent strategy addresses one of the key challenges in applying LLMs to healthcare: the need for domain-specific expertise. While LLMs possess broad knowledge, they may lack the specialized understanding required for specific medical tasks. By creating agents with defined roles, we can fine-tune each agent or utilize specific biomedical-pertained models as agents to align more closely with the specialized knowledge and decision-making processes of different ED professionals.

\subsection{Clinical Implications}

The CDSS's strong performance in clinical decision-making across various categories (primary diagnosis, critical findings, disposition decision, immediate action, medication, diagnostic tests, consultation, and monitoring) suggests its potential to enhance the quality and consistency of care. The system could help standardize care practices by providing accurate, evidence-based recommendations, potentially reducing variability in treatment decisions and improving adherence to clinical guidelines.

The system's pharmacological recommendations, supported by integration with the RxNorm database, generally resulted in medically correct suggestions and have implications for medication safety. By providing an additional layer of checking for drug interactions and appropriate dosing, the CDSS could help reduce medication errors, a significant concern in emergency care settings. However, we observed instances where the system recommended medicines that were not widely used or prohibitively expensive. This highlights the importance of contextualizing the CDSS with hospital-specific formularies and considering medication costs. In real-world applications, integrating the hospital's own medication database and providing the agent with context about medication pricing could significantly enhance the practical utility of these recommendations.

The system's comprehensive output, covering aspects from initial triage to disposition decisions, could facilitate better communication and handovers between different ED staff members. This improved information flow could lead to more coordinated care and potentially reduce errors related to miscommunication.

In terms of resource allocation, the CDSS's accurate predictions for necessary diagnostic tests, consultations, and monitoring could help optimize the use of hospital resources. This efficiency could be particularly beneficial in addressing the challenges of ED overcrowding, potentially reducing length of stay and improving patient throughput.

It's crucial to note that while the CDSS shows promise in enhancing various aspects of emergency care, it is designed to support, not replace, clinical decision-making. The system's recommendations should be used in conjunction with clinical expertise and judgment.

Importantly, the effective integration of the CDSS into clinical workflows remains a critical consideration. The system should complement rather than disrupt the established triage and treatment planning processes. Developing methodologies for seamless integration is essential for the system's success in real-world settings. This could involve creating an intuitive user interface (UI) that aligns with ED staff's needs and workflow patterns. Additionally, integration with the hospital's existing Electronic Medical Record (EMR) system could significantly enhance the CDSS's utility, allowing for real-time access to patient data and smoother incorporation of CDSS recommendations into patient care processes.

Furthermore, the clinical implications of this CDSS extend beyond immediate patient care. By providing a standardized approach to decision-making, the system could also serve as a valuable tool for training new ED clinicians, offering consistent, evidence-based guidance that can supplement experiential learning.

\subsection{Ethical Considerations}

The implementation of AI-driven clinical decision support systems in emergency medicine raises important ethical considerations and data privacy concerns that must be carefully addressed.

First and foremost, patient privacy and data protection are paramount. Our multi-agent CDSS was designed with these concerns in mind, utilizing the Llama-3-70b model, which can be run locally. This approach significantly reduces the risk of sensitive patient data being transmitted or stored externally, aligning with stringent healthcare data protection regulations such as HIPAA in the United States or GDPR in Europe\cite{b21,b22}. By keeping data processing on-premise, we mitigate the risks associated with data breaches or unauthorized access that could occur with cloud-based solutions.

However, the use of AI in clinical decision-making raises questions about accountability and transparency. While our system demonstrates high accuracy, it's crucial to maintain a "human-in-the-loop" approach where healthcare professionals retain ultimate responsibility for clinical decisions. The CDSS should be viewed as a tool to augment, not replace, clinical judgment. Clear guidelines must be established to delineate the roles of the AI system and human clinicians, ensuring that the ethical and legal responsibilities of care remain with healthcare professionals.

Transparency in AI decision-making processes is another critical ethical consideration. Although large language models like Llama-3-70b can provide highly accurate outputs, their decision-making processes can be opaque. This "black box" nature of AI can be problematic in healthcare settings where the rationale behind decisions is crucial. Future development should focus on improving the explainability of AI recommendations, possibly through the integration of explainable AI techniques.

Bias in AI systems is a significant ethical concern that requires ongoing attention \cite{b35}. While our multi-agent system aims to mimic the collaborative nature of ED teams, it's important to recognize that biases present in training data or model architecture could lead to systematic errors or unfair treatment of certain patient groups. Regular system performance audits across diverse patient populations will be necessary to identify and mitigate any biases.

The potential for AI systems to influence clinical decision-making also raises questions about patient autonomy and informed consent. Patients should be made aware when AI systems are being used to support their care, and consideration should be given to how this information is communicated without causing undue concern or eroding trust in healthcare providers.

\subsection{Limitations and Future work}

One significant limitation we encountered was related to the dataset used in the study. The Asclepius dataset, which is very descriptive and complex, may not fully represent real-world scenarios. Although comprehensive data is generally beneficial, real-world datasets in emergency medicine are often characterized by sparsity, imbalance, and a substantial proportion of missing values. These features pose significant challenges in developing robust models for clinical decision support. The discrepancy is particularly evident in emergency situations, such as those handled through emergency telephone calls. In these scenarios, information is often incomplete, rapidly changing, and collected under high-stress conditions. Emergency responders and dispatchers have been dealing with such data limitations since the inception of emergency call systems, often making critical decisions based on partial information and relying heavily on experience and standardized protocols to fill in the gaps. However, the Asclepius database, which we used to evaluate our CDSS, contains comprehensive and well-structured data. While this allowed for a thorough initial assessment of our system's capabilities, it also means that our evaluation benchmarks may differ from what we would expect in real-world performance. The system's ability to handle incomplete or inconsistent data, a crucial skill in actual emergency settings, was not fully tested in this study. This limitation underscores the need for future research to bridge the gap between idealized datasets and the messy reality of emergency medicine, incorporating more realistic data scenarios in subsequent evaluations of our CDSS to better assess its robustness and applicability in real-world emergency departments.

An important consideration in our research was the potential use of real-world emergency phone call data. Although we were unable to acquire such data for this study, our AI system has the potential to interpret emergency situations and provide crucial information to ED staff by integrating Speech-to-Text (STT) models. This capability could significantly enhance preparedness in the ED, allowing medical staff to anticipate incoming patients' needs more effectively. By providing early insights into the patient's condition, the system could facilitate the proactive preparation of resources such as treatment spaces, medications, diagnostic tests, and necessary medical equipment. This advance notice could streamline ED operations, potentially reducing wait times and improving the overall quality of emergency care by ensuring that appropriate resources are ready when the patient arrives.

While we equipped our AI with internet search capabilities via the DuckDuckGo tool, relying on internet searches for medical information is not optimal. The potential for information overload, exposure to false information, lack of source credibility, and the slight time delay associated with web searches could lead to misinformation, confusion, and inefficiencies in clinical decision-making. A more promising approach would be to utilize verified information from medical textbooks and databases, though this presents its own challenges in terms of copyright issues and credibility verification. To enhance the quality of our CDSS's responses, implementing standardized medical terminologies such as the Systematized Nomenclature of Medicine Clinical Terms (SNOMED CT) \cite{b32} and Logical Observation Identifiers Names and Codes (LOINC) \cite{b34} could be beneficial. Furthermore, integrating the CDSS into existing Electronic Medical Record (EMR) systems could significantly improve the productivity of ED physicians and nurses by assisting in the creation of medical records based on trained medical concepts. This integration would not only streamline workflow but also ensure that the AI's recommendations are grounded in verified, standardized medical knowledge, thereby enhancing the overall quality and reliability of the clinical decision support provided.

\section*{Conclusion}

In conclusion, our multi-agent CDSS represents a significant step forward in the application of AI to emergency medicine. By demonstrating high performance across the spectrum of emergency care tasks, our system shows promise in enhancing emergency care quality, efficiency, and safety. As we move forward with further validation and refinement, this technology has the potential to significantly improve patient outcomes in these critical healthcare settings.

\section*{Acknowledgments}

We would like to thank Sangkuk Han, MD, Ph.D., for reviewing the model and evaluating the answers. Wongyung Choi is the corresponding author.

\section*{Data Availability}

The data used in this study is publicly available through the Asclepius dataset, which can be accessed online.

\section*{Code Availability}

The code used for this research is openly available in a GitHub repository at \url{https://github.com/junhan51/LLM_ED_CDSS}.

\section*{Competing Interests}

Not applicable.

\vspace{12pt}

\newpage

\clearpage
\onecolumn

\renewcommand{\thesection}{A.\arabic{section}}
\renewcommand{\thefigure}{A.\arabic{figure}}
\renewcommand{\thetable}{A.\arabic{table}} 
\renewcommand{\theequation}{A.\arabic{equation}} 
\renewcommand{\theHsection}{A\arabic{section}}

\setcounter{section}{0}
\setcounter{figure}{0}
\setcounter{table}{0}
\setcounter{equation}{0}


\noindent \textbf{\LARGE{Appendix}}\\
\normalfont

In the following sections, we report additional data and detailed analyses to further illustrate our model.

We provide details on:
\begin{itemize}[]
\item Prompts used in LLM agents
\item Prompts used in LLM tasks
\item Example inputs and outputs

\end{itemize}

\vspace{0.8cm}

\section{Prompts used in LLM agents}

\subsection{Triage Nurse}

\begin{tcolorbox}
\textbf{Goal}: Conduct a thorough and rapid assessment of incoming patients using the KTAS (Korean Triage and Acuity Scale) system. Gather comprehensive information about patients' symptoms, vital signs, and relevant medical history to determine the urgency of care needed and facilitate efficient patient flow in the emergency department. \\

\textbf{Backstory}: You are a highly experienced triage nurse with over 15 years of experience in busy emergency departments. Your expertise in rapidly and accurately assessing patient conditions has been crucial in saving countless lives. You have a reputation for remaining calm under pressure and have trained numerous junior nurses in effective triage techniques.
\end{tcolorbox}

\vspace{0.8cm}

\subsection{Emergency Physician}

\begin{tcolorbox}
\textbf{Goal}: Provide rapid, accurate diagnoses and develop comprehensive treatment plans based on triage information and additional examinations. Make critical decisions about immediate interventions, prescribe appropriate medications, and coordinate with specialists when necessary to ensure optimal patient outcomes. \\

\textbf{Backstory}: As a board-certified emergency physician with over a decade of experience, you have handled a wide range of medical emergencies, from major traumas to complex medical conditions. Your ability to make quick, accurate diagnoses and implement effective treatment plans has earned you the respect of your colleagues and the trust of your patients. You are known for your calm demeanor in high-stress situations and your commitment to evidence-based medicine.
\end{tcolorbox}

\vspace{0.8cm}

\subsection{Pharmacist}

\begin{tcolorbox}
\textbf{Goal}: Ensure safe and effective medication use in the emergency room by leveraging 
        your extensive knowledge of pharmacology and the RxNorm database. Review 
        prescriptions meticulously, perform comprehensive drug interaction checks, 
        and provide crucial information about proper medication administration, 
        potential side effects, and dosage adjustments for emergency situations. \\

\textbf{Backstory}: You are a highly skilled pharmacist with specialized training in emergency 
        medicine. With over 15 years of experience in hospital pharmacy and emergency 
        departments, your expertise in managing complex medication regimens and 
        preventing adverse drug events is unparalleled. You have contributed to 
        developing hospital-wide protocols for medication safety in emergency 
        situations and are known for your ability to provide rapid, accurate 
        pharmacological consultations in high-pressure environments.
\end{tcolorbox}

\vspace{0.8cm}

\subsection{ED Doctor in Charge}

\begin{tcolorbox}
\textbf{Goal}: Oversee and coordinate all aspects of patient care in the emergency department. 
        Make critical clinical decisions regarding patient management, including 
        treatment plans, admission, and follow-up care. Ensure efficient utilization 
        of resources while maintaining the highest standards of patient care and safety. \\

\textbf{Backstory}: With over 20 years of experience in emergency medicine, including 10 years in 
        a leadership role, you are renowned for your clinical acumen and ability to 
        manage complex cases. Your expertise spans across all areas of emergency care, 
        and you have a track record of implementing innovative protocols that have 
        significantly improved patient outcomes and department efficiency. You are 
        respected for your decisive leadership in critical situations and your 
        commitment to mentoring junior staff.
\end{tcolorbox}

\vspace{0.8cm}
\section{Prompts Used in LLM Tasks}

\subsection{Medical Diagnosis}
\begin{tcolorbox}[breakable=true]
\textbf{Description}: 
        Based on the triage report and any additional examinations, provide a comprehensive 
        diagnosis and treatment plan for the patient with these symptoms:\\
        
        $\langle symptoms \rangle$ \\
        \{input\} \\
        $\langle /symptoms \rangle$ \\
        
        Your assessment should include: \\
        1. A thorough analysis of the patient's symptoms, vital signs, and medical history \\
        2. Differential diagnoses, listing the most likely conditions in order of probability \\
        3. Recommended diagnostic tests or imaging studies, with justification for each \\
        4. A detailed initial treatment plan, including medications, procedures, and interventions \\
        5. Consideration of potential complications and how to mitigate them \\
        6. Any necessary consultations with specialists, with clear reasons for each referral \\
        7. A plan for ongoing monitoring and reassessment of the patient's condition \\

        Use the search tool to find the latest evidence-based guidelines or unusual clinical presentations if necessary. \\

\textbf{Expected Output}:
        Provide a comprehensive medical report including: \\
        1. Primary working diagnosis with supporting evidence \\
        2. Differential diagnoses in order of likelihood, with brief explanations \\
        3. Detailed treatment plan, including: \\
           a. Medications (dosage, route, frequency) verified with RxNorm \\
           b. Immediate interventions or procedures \\
           c. Fluid management if applicable \\
           d. Pain management strategy \\
        4. Diagnostic tests ordered, with justification for each \\
        5. Potential complications to monitor for, with specific warning signs \\
        6. Consultations requested, if any, with clear rationale \\
        7. Plan for ongoing monitoring and criteria for reassessment \\
        8. Relevant evidence-based guidelines or literature referenced (if search tool was used) \\

        Format your response as follows: \\
        
        PRIMARY DIAGNOSIS: [State the most likely diagnosis] \\
        
        Supporting Evidence: [List key findings that support this diagnosis] \\
        
        Differential Diagnoses: \\
        1. [Diagnosis 1]: [Brief explanation] \\
        2. [Diagnosis 2]: [Brief explanation] \\
        3. [Diagnosis 3]: [Brief explanation] \\
        
        Treatment Plan: \\
        1. Medications: \\
           a. [Drug name, dosage, route, frequency] - [Purpose] (RxNorm verified) \\
           b. [Drug name, dosage, route, frequency] - [Purpose] (RxNorm verified) \\
        2. Interventions/Procedures: [List and describe] \\
        3. Fluid Management: [If applicable] \\
        4. Pain Management: [Strategy] \\
        
        Diagnostic Tests: \\
        1. [Test name]: [Justification] \\
        2. [Test name]: [Justification] \\
        
        Potential Complications: \\
        1. [Complication]: [Warning signs and management plan] \\
        2. [Complication]: [Warning signs and management plan] \\
        
        Consultations: \\
        1. [Specialist]: [Rationale for consultation] \\
        
        Monitoring Plan: [Describe ongoing monitoring and reassessment criteria] \\
        
        Evidence-Based Guidelines: [Reference any guidelines or literature used, if applicable]

\end{tcolorbox}

\subsection{Medication Review}
\begin{tcolorbox}[breakable=true]
\textbf{Description}: 
        Conduct a comprehensive review of the prescribed medications for the patient with these symptoms: \\
        
        $\langle symptoms \rangle$ \\
        \{input\} \\
        $\langle /symptoms \rangle$ \\

        Consider their medical history, current condition, and the emergency physician's treatment plan.  \\
        Your review should include: \\
        1. A thorough analysis of each prescribed medication using the RxNorm tool \\
        2. Potential drug interactions, including severity and clinical significance \\
        3. Dose appropriateness considering the patient's condition, age, weight, and renal/hepatic function \\
        4. Any contraindications based on the patient's medical history or current condition \\
        5. Identification of any high-alert medications that require special handling or monitoring \\
        6. Recommendations for medication adjustments, alternatives, or additional monitoring if necessary \\
        7. Important side effects or adverse reactions to watch for in the emergency setting \\
        8. Specific administration instructions or precautions for the nursing staff \\
        9. Any drug-disease interactions that could impact the patient's current condition \\

        Use the RxNorm tool extensively to gather detailed information about each medication. \\
        Use the search tool to find the latest pharmacological guidelines or information on rare
        drug effects if necessary. \\

\textbf{Expected Output}: 
        Provide a detailed medication safety report including: \\
        1. List of prescribed medications with a comprehensive analysis of each \\
        2. Identified drug interactions, contraindications, or concerns \\
        3. Dose appropriateness evaluations and any recommended adjustments \\
        4. Specific administration instructions and precautions \\
        5. Potential adverse effects to monitor in the emergency setting \\
        6. Recommendations for medication therapy optimization \\
        7. Any additional pharmacological considerations relevant to the patient's care \\

        Format your response as follows: \\
        
        MEDICATION SAFETY REPORT \\
        
        Patient Condition Summary: [Brief overview of relevant clinical information] \\
        
        Prescribed Medications Analysis: \\
        1. [Drug Name] (RxNorm verified): \\
           - Dose/Route/Frequency: [As prescribed] \\
           - Indication: [For what condition] \\
           - Appropriateness: [Comment on dose, considering patient factors] \\
           - Interactions: [List any significant interactions] \\
           - Contraindications: [If any, based on patient's condition] \\
           - Administration Instructions: [Specific guidance for nursing] \\
           - Monitoring: [What to watch for, including adverse effects] \\
        
        2. [Repeat for each medication] \\
        
        Overall Medication Therapy Assessment: \\
        - Drug-Disease Interactions: [Any concerns with patient's conditions] \\
        - High-Alert Medications: [Identify any that require special precautions] \\
        - Pharmacokinetic Considerations: [Any adjustments needed for renal/hepatic function] \\
        
        Recommendations: \\
        1. [Specific recommendation for medication adjustment, monitoring, or alternative] \\
        2. [Additional recommendations as needed] \\
        
        Emergency Pharmacology Considerations: \\
        - [Any special considerations for medication use in the ER setting] \\
        
        References: \\
        - [List any guidelines or resources used from the search tool, if applicable]

\end{tcolorbox}

\subsection{Triage Assessment}
\begin{tcolorbox}[breakable=true]
\textbf{Description}: 
        Conduct a comprehensive triage assessment of the patient presenting with these symptoms: \\

        $\langle symptoms \rangle$ \\
        \{input\} \\
        $\langle /symptoms \rangle$ \\
        
        Utilize the KTAS system to determine the urgency of care needed. Your assessment should include: \\
        1. A detailed evaluation of the patient's current symptoms and their severity \\
        2. Complete set of vital signs (blood pressure, heart rate, respiratory rate, temperature, oxygen saturation) \\
        3. Relevant medical history, including chronic conditions, allergies, and current medications \\
        4. Any recent trauma or significant events related to the current condition \\
        5. Pain assessment using a standardized scale \\
        6. Mental status evaluation \\
        7. Any immediate life-threatening conditions that require urgent intervention \\

        Refer to the KTAS (Korean Triage and Acuity Scale) guidelines below. \\
        $\langle KTAS \: guide \rangle$ \\
        1:  \\
            description: "Conditions that require immediate intervention; life-threatening or potentially life-threatening states (or high risk of rapid deterioration)", \\
            examples: ["Cardiac arrest", "Respiratory arrest", "Unconsciousness not related to alcohol consumption"], \\
            priority: "Highest priority" \\
        
        2:  \\
            description: "Conditions with potential threats to life, limb, or organ function requiring rapid medical intervention", \\
            examples: ["Myocardial infarction", "Cerebral hemorrhage", "Cerebral infarction"], \\
            priority: "Second priority" \\
        
        3:  \\
            description: "Conditions that may progress to a serious problem requiring emergency intervention", \\
            examples: ["Dyspnea (with oxygen saturation above 90\%)", "Diarrhea with bleeding"], \\
            priority: "Third priority" \\
        
        4:  \\
            description: "Conditions that, considering the patient's age, pain level, or potential for deterioration or complications, could be treated or re-evaluated within 1-2 hours", \\
            examples: ["Gastroenteritis with fever above 38°C", "Urinary tract infection with abdominal pain"], \\
            priority: "Fourth priority" \\
        
        5:  \\
            description: "Conditions that are urgent but not emergencies, or those resulting from chronic problems with low risk of deterioration", \\
            examples: ["Common cold", "Gastroenteritis", "Diarrhea", "Laceration (wound)"], \\
            priority: "Fifth priority" \\
        
        $\langle /KTAS \: guide \rangle$ \\

        Use the search tool to find any additional information about unusual symptoms or conditions if necessary. \\

\textbf{Expected Output}: 
        Provide a comprehensive triage report including: \\
        1. KTAS level assigned (1-5) with detailed justification based on the KTAS criteria \\
        2. Comprehensive summary of symptoms, vital signs, and relevant medical history \\
        3. List of current medications and allergies, verified using the RxNorm tool \\
        4. Any critical or unusual findings that require immediate attention \\
        5. Recommended immediate actions or precautions for the medical team \\
        6. Any additional information gathered from the search tool, if used \\

        Format your response as follows: \\
        
        KTAS CLASSIFICATION: [Level] \\
        
        Detailed Justification: [Explain why this KTAS level was assigned] \\
        
        Patient Assessment: \\
        1. Presenting Symptoms: [List and describe] \\
        2. Vital Signs: [List all measured vital signs] \\
        3. Medical History: [Relevant past and current conditions] \\
        4. Medications and Allergies: [List with RxNorm verifications] \\
        5. Pain Assessment: [Score and description] \\
        6. Mental Status: [Brief evaluation] \\
        
        Critical Findings: [Any immediate life-threatening conditions] \\
        
        Recommended Actions: [Immediate steps for the medical team] \\
        
        Additional Information: [Any relevant data from search tool, if used] \\

\end{tcolorbox}

\subsection{ED Management Decision}
\begin{tcolorbox}[breakable=true]
\textbf{Description}: 
        As the Emergency Room Doctor in Charge, review all the information provided by the triage nurse, 
        emergency physician, and pharmacist for the patient with these symptoms: \\
        
        $\langle symptoms \rangle$ \\
        \{input\} \\
        $\langle /symptoms \rangle$ \\
        
        Based on this comprehensive 
        assessment, make critical clinical decisions regarding the patient's care. Your decision-making 
        process should include: \\
        1. A thorough review of the KTAS classification and its implications for immediate care \\
        2. Evaluation of the diagnosis, differential diagnoses, and proposed treatment plan \\
        3. Consideration of the medication safety report and any pharmacological concerns \\
        4. Assessment of the need for immediate interventions, further diagnostic tests, or specialist consultations \\
        5. Determination of the most appropriate next steps for patient care (e.g., continued ER management, 
           admission to a specific unit, transfer to a specialized facility, or safe discharge with follow-up) \\
        6. Consideration of resource utilization and department capacity in your decision-making \\
        7. Development of a clear, actionable plan for ongoing patient care and monitoring \\

        Use the RxNorm tool to double-check any medication decisions. Use the search tool to find relevant 
        clinical guidelines or hospital protocols if necessary for decision-making. \\

\textbf{Expected Output}:
        Provide a comprehensive management decision including: \\
        1. Restatement of the KTAS classification with your assessment of its accuracy \\
        2. Your agreement or adjustments to the primary diagnosis and treatment plan \\
        3. Critical decision on patient disposition (continue ER care, admit, transfer, or discharge) \\
        4. Detailed justification for your decision, including clinical and logistical factors \\
        5. Specific instructions for next steps in patient care, including any changes to the treatment plan \\
        6. Additional resources, specialists, or interventions needed \\
        7. Clear communication plan for the patient, family, and other healthcare providers \\
        8. Contingency plans or criteria for reassessing the patient's condition \\

        Format your response as follows: \\
        
        EMERGENCY DEPARTMENT MANAGEMENT DECISION \\
        
        KTAS Classification Review: [Restate and assess accuracy] \\
        
        Clinical Assessment: \\
        - Primary Diagnosis: [State agreement or provide alternative with justification] \\
        - Critical Findings: [Highlight key issues requiring immediate attention] \\
        
        Disposition Decision: [State clearly: Continue ER care / Admit / Transfer / Discharge] \\
        
        Justification: [Provide a detailed explanation for your decision, referencing clinical findings, resource considerations, and best practices]
        
        Management Plan: \\
        1. Immediate Actions: [List any immediate interventions required] \\
        2. Medications: [Any changes or confirmations to the proposed medication plan] \\
        3. Diagnostic Tests: [Additional tests ordered or results to be followed up] \\
        4. Consultations: [Any specialist consultations required, with urgency level] \\
        5. Monitoring: [Specific parameters to be monitored and frequency] \\
        
        Resource Allocation: [Specify any special resources needed and how they will be utilized] \\
        
        Communication Plan: \\
        - Patient/Family: [Key points to be communicated] \\
        - Healthcare Team: [Instructions for ER staff, admitting team, or follow-up providers] \\
        
        Contingency Planning: [Criteria for reassessment or change in management plan] \\
        
        Additional Considerations: [Any other relevant factors influencing your decision] \\
        
        References: [List any guidelines or protocols referenced, if search tool was used] \\

\end{tcolorbox}

\vspace{0.8cm}
\section{Example Input and Output}

\begin{tcolorbox}[breakable=true]
\textbf{Input} \\

The patient was hospitalized due to persistent fever and dry cough for 2 weeks, leading to reduced general health condition. On admission, the patient required 4 L/min of oxygen due to rapid shallow breathing at rest and severe breathlessness during minor physical activity. The initial physical therapy focused on educating the patient on dyspnea-relieving positions, mobilization techniques, and deep-breathing exercises. However, with time, the patient's anxiety-induced dyspnea became an issue, leading to the modification of therapy to relieve his dyspnea.
\end{tcolorbox}

\begin{tcolorbox}[breakable=true]
\textbf{Output} \\

EMERGENCY DEPARTMENT MANAGEMENT DECISION \\

KTAS Classification Review: The patient's symptoms and vital signs indicate a KTAS classification of 3, which is accurate given the severity of their respiratory distress and potential for rapid deterioration. \\

Clinical Assessment: \\
- Primary Diagnosis: Acute Respiratory Distress Syndrome (ARDS) with underlying Pneumonia \\
- Critical Findings: The patient's respiratory compromise, anxiety-induced dyspnea, and potential underlying medical conditions require immediate attention. \\

Disposition Decision: Admit to the Intensive Care Unit (ICU) for close monitoring and management of respiratory distress, pneumonia, and potential complications. \\

Justification: \\
The patient's condition is critical and requires immediate attention. Admission to the ICU will provide close monitoring and management of their respiratory distress, pneumonia, and potential complications. The ICU team will be able to provide oxygen therapy, mechanical ventilation, and other supportive therapies as needed. \\

Management Plan: \\

1. Immediate Actions: \\
    - a. Continue oxygen therapy at 4 L/min via nasal cannula \\
    - b. Administer antibiotics (Ceftriaxone) and bronchodilators (Albuterol) as prescribed \\
    - c. Provide anxiolytics (Lorazepam) as needed to relieve anxiety-induced dyspnea \\
    - d. Perform chest X-ray and arterial blood gas (ABG) analysis to evaluate for pneumonia and respiratory complications \\
2. Medications: \\
    - a. Oxygen therapy: 4 L/min via nasal cannula \\
    - b. Antibiotics: Ceftriaxone 1g IV q12h \\
    - c. Bronchodilators: Albuterol 2.5mg nebulized q4h \\
    - d. Anxiolytics: Lorazepam 1mg IV q6h as needed \\
3. Diagnostic Tests: \\
    - a. Chest X-ray \\
    - b. Arterial blood gas (ABG) analysis \\
    - c. Complete blood count (CBC) \\
    - d. Blood cultures \\
4. Consultations: \\
    - a. Pulmonology: To evaluate and manage respiratory failure and ARDS \\
    - b. Infectious disease: To evaluate and manage pneumonia or sepsis \\
5. Monitoring: \\
    - a. Continuously monitor oxygen saturation, respiratory rate, and heart rate \\
    - b. Perform regular ABG analysis to monitor oxygenation and acid-base status \\
    - c. Evaluate for signs of respiratory failure, sepsis, or cardiac complications \\

Resource Allocation: \\
The ICU team will provide close monitoring and management of the patient's condition. Additional resources, such as respiratory therapy and pharmacy, will be utilized as needed. \\

Communication Plan: \\
- Patient/Family: The patient's condition and treatment plan will be explained, and they will be encouraged to ask questions and express concerns. \\
- Healthcare Team: The ICU team will be informed of the patient's condition and treatment plan, and will be responsible for providing close monitoring and management. \\

Contingency Planning: \\
The patient's condition will be reassessed regularly, and the treatment plan will be adjusted as needed. If the patient's condition deteriorates, consideration will be given to intubation and mechanical ventilation. \\

Additional Considerations: \\
The patient's anxiety-induced dyspnea will be addressed through the use of anxiolytics and other supportive therapies. The patient's medications will be monitored for potential interactions and adverse effects. \\

References: \\
- American Thoracic Society (ATS) guidelines for the management of ARDS \\
- Infectious Diseases Society of America (IDSA) guidelines for the management of pneumonia \\

\end{tcolorbox}

\newpage

\clearpage

\end{document}